\documentclass[letterpaper, 10 pt, conference]{ieeeconf}  % Comment this line out if you need a4paper

\IEEEoverridecommandlockouts                              % This command is only needed if 
\usepackage{amsmath} % assumes amsmath package installed
\usepackage{subfigure}

\usepackage{todonotes}
\usepackage{amsmath}
\usepackage{amssymb}
\usepackage{diagbox}
\usepackage{makecell}
\usepackage{amsmath}
\usepackage{url}
\usepackage{hyperref}
\usepackage{bm}
\usepackage{array}
\let\labelindent\relax
\usepackage{enumitem}
\usepackage{color,soul}
\usepackage{leftindex}
\usepackage{graphicx}
\usepackage{tensor}   % For tensor notation with better spacing
\usepackage{stackengine}

\usepackage{adjustbox}
\usepackage{threeparttable}
\usepackage{multirow}
\usepackage{booktabs}

\title{\LARGE \bf
A Self-Supervised Learning Approach with Differentiable Optimization for UAV Trajectory Planning
}
\author{Yufei Jiang$^{*1}$, Yuanzhu Zhan$^{*1}$, Harsh Vardhan Gupta$^{2}$, Chinmay Borde$^{2}$, Junyi Geng$^{1}$%
\thanks{$^{*}$ Equal contribution in alphabetical order.}%
\thanks{$^{1}$ Department of Aerospace Engineering, Pennsylvania State University, University Park, PA, 16802, USA. 
{\tt\footnotesize \{yufei\_jiang, yvz6008, jgeng\}@psu.edu}}%
\thanks{$^{2}$ Department of Engineering and Applied Science, University at Buffalo, NY, 14260, USA.
{\tt\footnotesize \{hgupta4, cborde\}@buffalo.edu}}%}% 
}

\begin{document}

\maketitle
% \thispagestyle{empty}
% \pagestyle{empty}

%%%%%%%%%%%%%%%%%%%%%%%%%%%%%%%%%%%%%%%%%%%%%%%%%%%%%%%%%%%%%%%%%%%%%%%%%%%%%%%%
\begin{abstract}
While Unmanned Aerial Vehicles (UAVs) have gained significant traction across various fields, path planning in 3D environments remains a critical challenge, particularly under size, weight, and power (SWAP) constraints. Traditional modular planning systems often introduce latency and suboptimal performance due to limited information sharing and local minima issues. End-to-end learning approaches streamline the pipeline by mapping sensory observations directly to actions but require large-scale datasets, face significant sim-to-real gaps, or lack dynamical feasibility. In this paper, we propose a self-supervised UAV trajectory planning pipeline that integrates a learning-based depth perception with differentiable trajectory optimization. A 3D cost map guides UAV behavior without expert demonstrations or human labels. Additionally, we incorporate a neural network-based time allocation strategy to improve the efficiency and optimality. The system thus combines robust learning-based perception with reliable physics-based optimization for improved generalizability and interpretability. Both simulation and real-world experiments validate our approach across various environments, demonstrating its effectiveness and robustness. Our method achieves a 30.90\% reduction in control effort while maintaining competitive tracking performance compared with state-of-the-art.
\end{abstract}
%%%%%%%%%%%%%%%%%%%%%%%%%%%%%%%%%%%%%%%%%%%%%%%%%%%%%%%%%%%%%%%%%%%%%%%%%%%%%%%%

\section{Introduction}
\label{sec: intro}

% Aerial robots and aerial path planning challenge -> tradition modular-based approach -> existing end-to-end learning -> Our proposed approach -> contribution summary

% Aerial robots and path planning challenge
Over the past decade, interest in Unmanned Aerial Vehicles (UAVs) has grown rapidly across a wide range of fields, including applications such as 3D mapping~\cite{zhao2021super}, exploration~\cite{hu2023off}, physical interaction~\cite{he2023image,guo2024flying} and package delivery~\cite{geng2020cooperative}. One of the critical UAV tasks is efficient path planning. In the environment without a pre-established map, the UAVs must re-plan quickly and efficiently to avoid collisions and perform safe navigation even under size, weight, and power (SWAP) constraints. 
Path planning for UAVs poses unique challenges comparing to the planning for ground mobile robot. First, UAVs operate in three-dimensional (3D) space, significantly increasing the search area and requiring consideration of obstacles at varying altitudes. Second, UAVs typically move at higher speeds, imposing stricter safety requirements. Third, planned paths must be dynamically feasible to account for 3D maneuvers, avoiding unplanned collisions resulting from discrepancies between planned and executed paths.

\begin{figure}
    \centering
    \includegraphics[width = 1\linewidth]{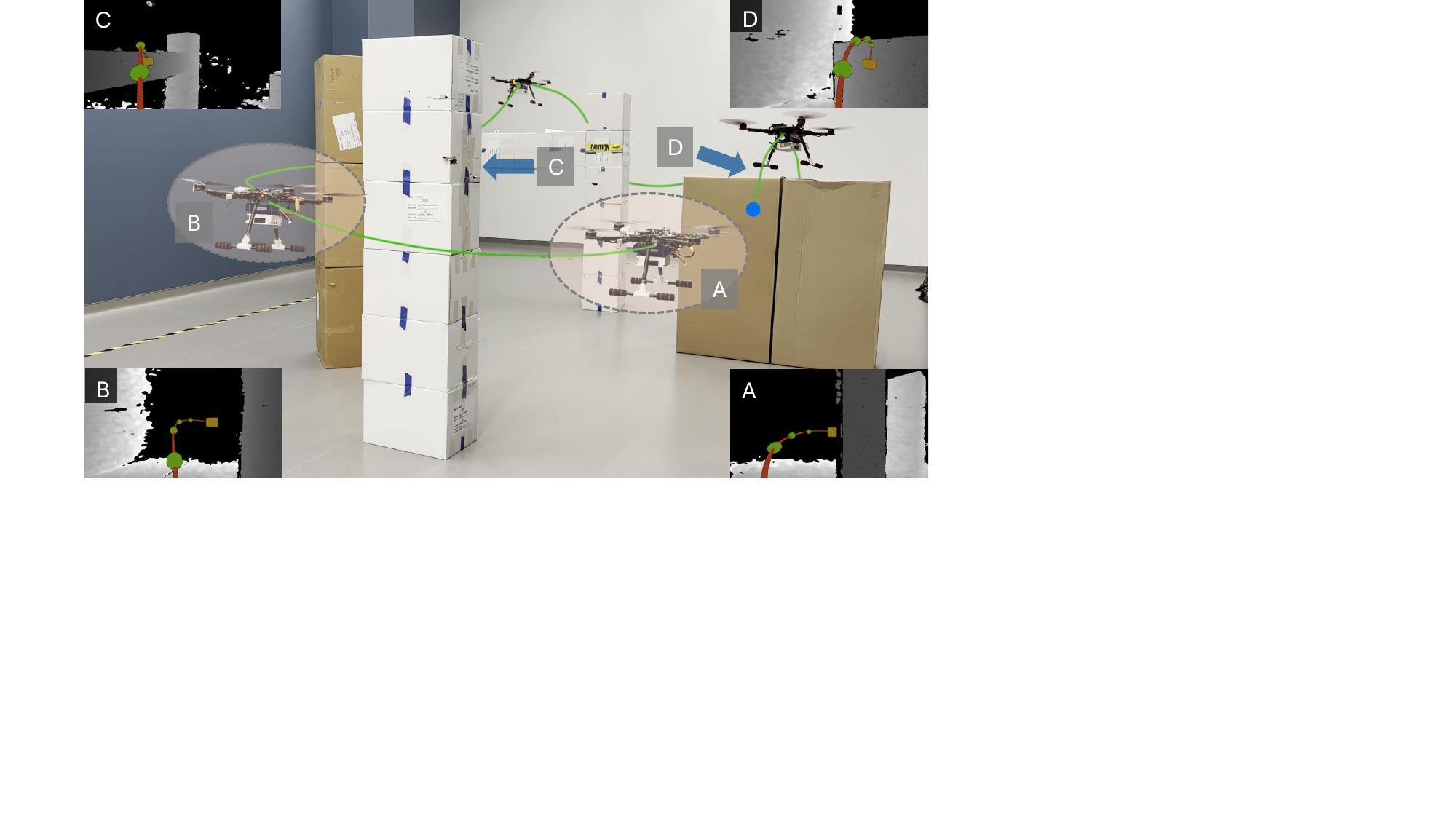}
    \caption{3D UAV trajectory planning in complex environment. Different desired waypoints are given by user, the UAV relies solely on depth images as input. The green curve showcases the UAV's trajectory. Starting from point (A) avoids vertical pillar (B), then performs maneuver to position (C) avoids the horizontal beam, (D) performs another vertical avoidance maneuver at different height and reach the blue target point.}
    \label{fig: concept}
    \vspace{-1.5em}
\end{figure}

% Traditonal modular approach: latency, limited by each module -> local minimum; iterative design
Traditional planning pipelines are typically modular, with perception, mapping, and path-searching modules designed separately and then integrated~\cite{cao2021tare, best2022resilient, cao2022autonomous}. This approach, however, has several drawbacks. Limited information sharing between modules introduces latency, slowing the system's response. Independent module design also leads to suboptimal performance, as conservative assumptions in one component may trap the system in local minima. Moreover, the modular separation requires iterative updates during testing, where parameters of one module are fine-tuned while others are frozon, making the process tedious and time-consuming.
% Exsiting end-to-end learning: SL: need large amount of data;  RL: low sample efficiency, poor sim-to-real gap
% IL based planning: only 2D, no iterative optimization, not applicable to drones
End-to-end learning, which map sensory observations directly to actions or trajectories, have gained attention for their streamlined pipelines and robust perception capabilities. However, they often require large datasets, suffer from low sample efficiency, and face significant sim-to-real gaps. These methods also lack interpretability, making it difficult to incorporate physical knowledge. For instance, some supervised learning approaches that imitate expert-generated trajectories perform well in controlled settings but struggle to generalize~\cite{loquercio2021learning}. Reinforcement learning (RL) and its variants aim to improve efficiency but often encounter issues like slow convergence, low sample efficiency, and overfitting~\cite{kulkarni2024reinforcement, kahn2021badgr}. Additionally, designing effective reward functions is time-consuming, and even with well-crafted rewards, sim-to-real transfer remains a major challenge.

Recent research has explored hybrid approaches that combine learning-based and model-based methods by integrating differentiable traditional components with learning modules \cite{wang2024imperative}. For example, \cite{Yang-RSS-23} introduced a self-supervised method combining a learning-based depth perception module with trajectory optimization (TO), using a pre-built traversability cost map for local planning. However, this approach is limited to ground robots due to its reliance on a 2D cost map projection. Moreover, it employs a closed-form solution for cubic spline in TO without gradient backpropagation as claimed Bi-level optimization (BLO), reducing its generalizability to more complex or constrained scenarios. 
While \cite{li2024ikap} extends \cite{Yang-RSS-23} with differentiable model predictive control (MPC) to enforce kinematic constraints, it guarantees only kinematic (not dynamic) feasibility due to its bicycle-model state choice. It is also limited to a 2D setting and handles only equality constraints.
%\cite{li2024ikap} also proposed an end-to-end planning framework, which integrates MPC for trajectory optimization. However, it can only guarantee the kinematics feasibility rather than dynamics feasibility due to their states selection for the bicycles model. 
\cite{han_gaofei2025RAL} embeds safe corridors and differentiable TO in the network for generating UAV dynamically feasible trajectories. However, it requires an offline motion-primitive library and still relies on supervised training. 
%\cite{Mellinger2011MinSnap} proposed a well-known trajectory generation method for quadrotors that minimizes control effort. However, time allocation remains a complex problem that has been widely studied. Iterative optimization via gradient descent \cite{Burri2015timealloc, Bry2015timealloc} is computationally expensive and impractical for real-time applications, while time parametrization methods often fail to guarantee optimality. 
%To address these challenges, we introduce a time allocation network that predicts the time for each trajectory segment, improving both efficiency and optimality compared to traditional approaches. Notably, this network is differentiable and can be updated through backpropagation based on the final trajectory loss, enabling it to learn an optimal time allocation strategy.

To address these limitation, we propose a self-supervised pipeline for UAV path planning that combines a learning-based depth perception with differentiable, metric-based TO, forming a bi-level optimization (BLO). Our approach includes a 3D cost map to account for UAVs' 3D operations, providing collision costs to guide behavior without requiring expert demonstrations or human labels. We also develop a differentiable minimum snap TO module to ensure dynamically feasible trajectories with both equality and inequality constraints and enable gradient backpropagation on iterative optimization for end-to-end training. Additionally, a time allocation network predicts segment durations, enhancing efficiency and optimality.
During deployment, the policy predicts collision conditions directly from first-person view (FPV) depth observations and plans paths accordingly. The end-to-end design optimizes observation features for planning objectives, combining the robustness of learning-based perception with the reliability, generalizability, and interpretability of physics-based methods.
%To account for UAVs' 3D operations, we develop a 3D cost map that provides traversability costs to guide UAV behavior, enabling the entire pipeline to function end-to-end without the need for expert demonstrations or human labels. To ensure trajectory feasibility, we developed a differentiable minimum snap trajectory optimization module, which generates dynamically feasible trajectories while allowing proper gradient backpropagation during end-to-end training. We also introduce a time allocation network that predicts the time for each trajectory segment to improve both efficiency and optimality.

%we developed a differentiable minimum snap trajectory optimization module with a neural network-based time allocation strategy, which addresses the inefficiency of iterative methods and the suboptimality of polynomial time allocation while allowing proper gradient backpropagation during end-to-end training. 

% contribution summary 
In summary, the main contributions of this work are:
\begin{itemize}[leftmargin=*]
    \item We create a self-supervised pipeline for 3D UAV path planning that combines a learning-based depth perception module with differentiable, metric-based TO. 
	
    \item We train the planner using geometry-derived collision signals from a 3D cost map for self-supervision without the need for expert demonstrations or human labels. %We introduce a 3D cost map that provides collision costs to guide UAV behavior for self-supervision without the need for expert demonstrations or human labels.

    \item We develop a differentiable minimum snap TO module for dynamically feasible trajectories while allowing end-to-end training. A time allocation network is designed to enhance efficiency and optimality.

    %\item We introduce a neural network-based time allocation strategy, which addresses the inefficiency of iterative methods and the suboptimality of polynomial time allocation.
    
    \item We present both simulation and real-world experiments to evaluate the proposed system in various environments.
\end{itemize}

\vspace{-1mm}
\section{Related Works}
\label{sec: related works}

\subsection{UAV Path Planning}
Many classic planning frameworks have demonstrated effective for autonomous UAV navigation in challenging environments. For example, the modular approach~\cite{cao2022autonomous} demonstrated in the DARPA Subterranean Challenge introduces latency and lacks robustness to noise and real-world errors. Gradient-based planning approaches optimize trajectories with manually designed safety constraints but can get stuck in local minima~\cite{zhou2021raptor, zhou2020ego, gao2020teach}. \cite{FastPlanner} is a UAV motion planning framework that leverages Euclidean Distance Fields (EDF) for gradient-based B-spline optimization. Similarly, corridor-based methods rely on low-dimensional search algorithms, often neglecting higher-order kinematics and producing dynamically infeasible trajectories~\cite{gao2020teach, tordesillas2021mader}. 
End-to-end learning methods have become emerging by eliminating explicit mapping and reducing latency. For instance, \cite{loquercio2021learning} trained a deep neural network (DNN) using human flight data, but it requires large datasets and struggles to generalize to diverse environments, with performance limited by the suboptimality of expert trajectories.
Reinforcement learning has also been explored for UAV path planning~\cite{kulkarni2024reinforcement}, using real-world data to overcome the sim-to-real gap~\cite{kahn2021badgr}. However, these approaches rely on neural networks and lack guarantees for kinematic feasibility and trajectory optimality. 

Recently, hybrid methods combining neural networks with numerical optimization are emerging. While many work still treated the two separately~\cite{jacquet2024n}, there are some research incorporates differentiable optimization for end-to-end training~\cite{han_gaofei2025RAL, wu2024dnntimeallocation, lu2024yopo}. For example, \cite{lu2024yopo} integrates perception and trajectory optimization by predicting the offsets and scores of a set of motion primitives for local optimization. However, the sampling-based method and predicted end-derivative cannot guarantee optimality. \cite{wu2024dnntimeallocation} integrate DNNs within the motion planning framework to predict the time allocation of a given set of waypoints. However, it formulates path searching as a separate module and rely on supervised learning for training. 
In contrast, our method avoids handcrafted primitives and uses self-supervision to jointly optimize perception and planning for 3D UAV scenario, reducing labeling and integration overhead.
%%However, \cite{yang2023iplanner} uses a closed-from solution for Cubic spline in the TO without gradient backpropagation as claimed BLO, reducing its generalizability to other scenarios with constraints. \cite{han2024learning} relies on supervised learning with results limited to simulation.
\vspace{-1mm}

\subsection{Differentiable Optimization}
%Optimization algorithms is a fundamental method in robotic planning, which can enforce equality and inequality constraints, ensuring feasibility under various physical and operational limitations while improving computational efficiency. However, pure optimization methods rely on accurate models and struggle with non-convex problems. To address these challenges, research has increasingly integrated optimization with learning. \cite{amos2017optnet} proposed OptNet and released qpth, embedding optimization as a differentiable layer within neural networks to enable gradient backpropagation for argmin problems. \cite{holmesSDPRLayers2024} extended this approach to semidefinite program relaxations (SDPR) for non-convex polynomial problems. Furthermore, \cite{cvxpylayers2019} introduced differentiable optimization layers for disciplined convex programs, representing them in an affine-solver-affine form to facilitate end-to-end differentiation, with applications in linear machine learning models and stochastic control. More recently, \cite{NEURIPS2024_8db12f72} proposed BPQP, which reformulates the backward pass as a quadratic programming problem, improving computational efficiency and enabling faster gradient computations compared to existing differentiable optimization layers.

Differentiable optimization refers to optimization algorithms where the output is differentiable with respect to the input parameters, enabling the integration of traditional model-based optimization with learning-based approaches. This paradigm combines the strengths of both methods—infusing interpretable domain knowledge into deep learning pipelines and reducing the need for tedious parameter tuning in traditional optimization. It has become a powerful paradigm in robotics for perception~\cite{teed2021droid}, planning~\cite{Yang-RSS-23}, control~\cite{jin2020pontryagin}, and physics-based simulation~\cite{zhang2024back}. A pioneering work, OptNet~\cite{amos2017optnet}, embeds optimization as a differentiable layer within neural networks, enabling gradient backpropagation for argmin problems. This approach has later been extended to model predictive control (MPC) using convex quadratic approximations for non-convex MPC problems~\cite{amos2018differentiable}, and to actor-critic MPC by integrating RL~\cite{romero2024actor}. 
% SDPRLayers further expanded this to semidefinite program relaxations for non-convex polynomial problems~\cite{holmesSDPRLayers2024}.
Recently, BPQP reformulated the backward pass as a quadratic programming (QP) problem~\cite{pan2025bpqp}, improving computational efficiency. Tools like Theseus~\cite{pineda2022theseus} and PyPose~\cite{wang2023pypose} have also emerged, offering differentiable optimization capabilities for robotics. Our approach formulates the problem as a BLO where gradient backpropagation from the upper to lower level is critical, and employs a differentiable QP solver to incorporate both equality and inequality physical constraints.

%\subsection{3D Cost Map}
\section{Methodology}
\label{sec: method}
\begin{figure*}
    \centering
    \includegraphics[width = 0.95\linewidth]{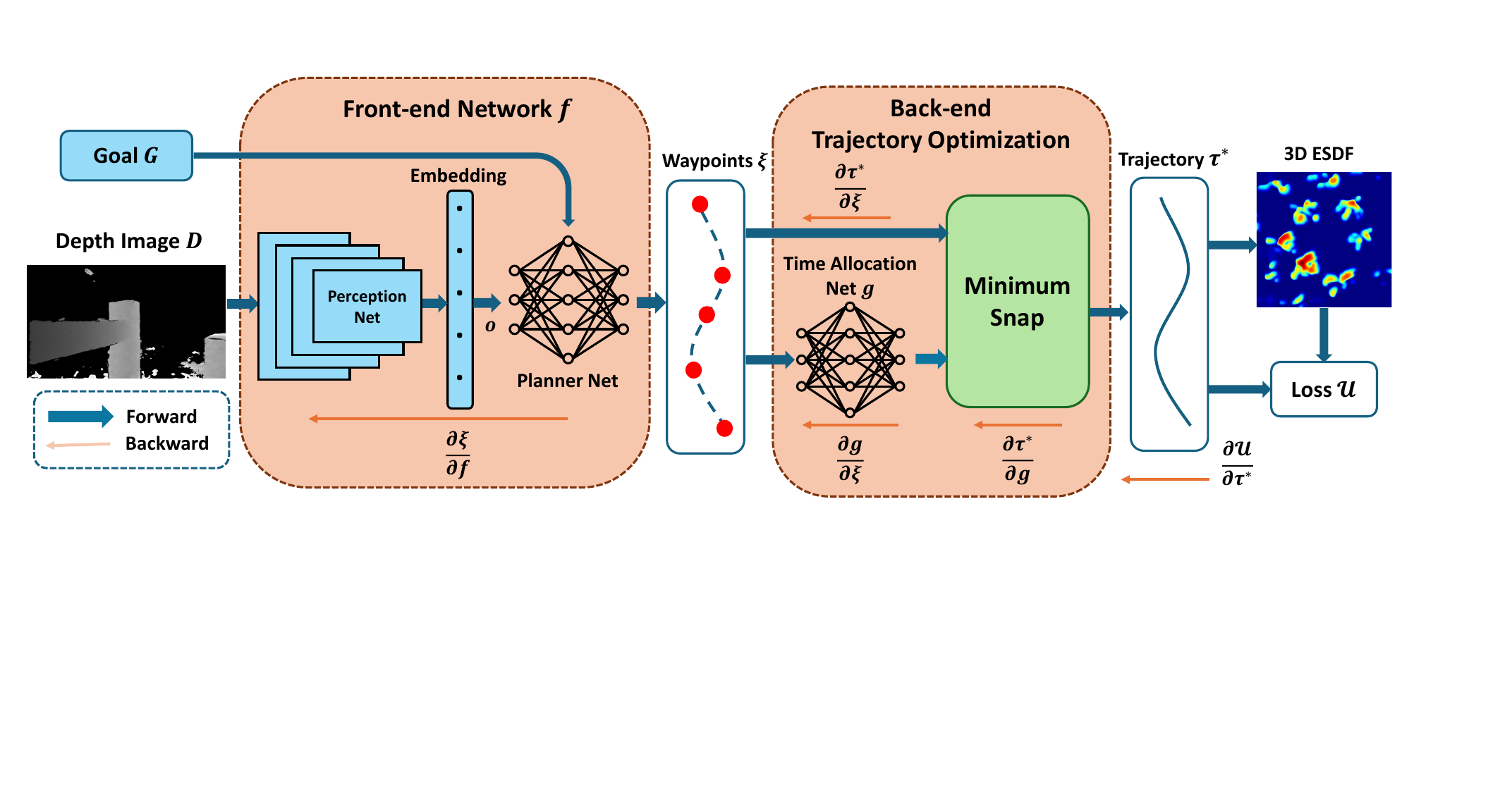}
    \caption{Overview of the planning pipeline. It consists of two parts, forming a bi-level optimization. The perception and planning network encodes the depth measurements and goal position to predict a key-point path with an collision probability. Then, the low-level trajectory optimization refines the path under specific constraints and cost. A well designed upper-level loss including trajectory cost and time allocation loss updates the network via gradients backpropagated through the trajectory optimizer.}
    \label{fig: pipeline}
    \vspace{-2em}
\end{figure*}

\subsection{Problem Definition}
At each time stamp, given the observation of the depth image $\mathcal{D}$ and a target location $_\mathcal{B}\mathbf{G}\in \mathbb{R}^3$ in the UAV body frame $\mathcal{B}$, the planning problem is to find a trajectory $\boldsymbol{\tau}$ to guide the UAV from the current position $\mathbf{p}^R \in \mathbb{R}^{3}$ to the target while avoiding obstacles $\mathcal{O} \subset \mathcal{M} \subset \mathbb{R}^3$, where $\mathcal{M}$ is the whole workspace, $\mathcal{O}$ is the obstacle space the UAV cannot fly through.

\subsection{Planning with Hybrid Learning and Optimization}

The proposed hybrid learning and optimization process for UAV planning is shown in Figure~\ref{fig: pipeline}. First, front-end network $f$: a convolutional neural network (CNN) front-end takes the depth input and encode into observation embedding. This embedding, combined with the target, is fed into the second planning network to predict a n-key-point path $\boldsymbol{\xi}\in\mathbb{R}^{n\times 3}$. In the back end, a Time Allocation Net $g$ takes into the key-point path and generates the timestamp $\mathcal{T} \in \mathbb{R}^n$ for each trajectory segment. Next, a differentiable minimum snap trajectory optimizer (MSTO) takes into both the key-point path, generated time allocation and outputs a dynamic feasible trajectory $\boldsymbol{\tau}^*$. Then, a well-designed collision costs $\mathcal{U}$ based on a 3D cost map is evaluated and backpropagated through both the differentiable MSTO and the networks to update the network parameters of both the front-end network $\boldsymbol{\theta}_f$ and the Time Allocation Net $\boldsymbol{\theta}_T$. This process results in a nested bi-level optimization (BLO) problem, where the networks and differentiable MSTO can be jointly optimized.
\begin{align}\label{pro:blo}
    \min_{\boldsymbol{\theta}_f, \boldsymbol{\theta}_T} &\ \mathcal{U}(f({\boldsymbol{\theta}_f}), g({\boldsymbol{\theta}_T}) , \boldsymbol{\tau}^*) \\
    \text{s.t.} &\ \boldsymbol{\tau}^* = \text{arg} \min_{\tau} \mathcal{L}(f({\boldsymbol{\theta}_f}), g({\boldsymbol{\theta}_T})) \nonumber\\
    &\quad\ \text{s.t.}\ h(f({\boldsymbol{\theta}_f}), g({\boldsymbol{\theta}_T})) \leq 0 \nonumber
\end{align}
where $\mathcal{L}$ represents the cost of differentiable MSTO, $h$ represents the constraints of the sub-optimization problem.

\subsection{Neural Networks}
\subsubsection{Perception Network}
Here we use a similar network structure as in~\cite{Yang-RSS-23}. At each time step, upon receiving a depth measurement $\mathcal{D}$, the perception network encodes this observation into a high-dimensional embedding $\mathbf{o}$ using ResNet-18 , a widely adopted and efficient CNN architecture to extract features while preserving spatial and geometric information for planning \cite{He2015DeepRL}. 

\subsubsection{Planning Network}
The planning network takes into the perception embedding $\mathbf{o}$ and the goal position $\mathbf{G}$ to compute a collision-free key-point path $\boldsymbol{\xi}$. The goal position is first mapped to a higher-dimensional feature embedding using a linear layer and then concatenated with the perception embedding to serve as the input for the planning network. The planning network consists of a CNN and MLP layers with ReLU activation to predict a key-point path $\boldsymbol{\xi}$ and the collision probability $\eta$. This path is then optimized by the following differentiable MSTO.

% \subsubsection{Time Allocation Network}
% Time Allocation Net is designed to predict the time allocation for each segment of a trajectory. It's taking the key-point path $\xi$ as input and allocated time $\mathcal{T} \in \mathbb{R}^n$ as output. To train it, we use a supervised learning approach, where the ground truth time allocation $\mathcal{T}^*$ is obtained through an iterative optimization process. This design choice addresses two key challenges in trajectory optimization. First, it mitigates the issue of poor initial time allocation, which can lead to the failure of minimum snap trajectory generation. Second, during inference, it eliminates the need for computationally expensive iterative time allocation, thereby significantly improving real-time performance. By leveraging data-driven learning, it provides an efficient and robust solution for time allocation in motion planning tasks.

\subsection{Differentiable Minimum Snap Trajectory Optimization (MSTO)}\label{sec:diff-opt}
To enforce the dynamic feasibility and ensure the safety of the local path generated by the planner, we develop a MSTO to optimize the predicted key-point path.
\subsubsection{UAV model}
The UAV dynamics can be modeled as the well-known Netwon-Euler method:
% \vspace{-3mm}
\setlength{\abovedisplayskip}{4pt}
\setlength{\belowdisplayskip}{4pt}
\begin{align}
    m\Ddot{\mathbf{r}} &= -mg\mathbf{z}_W + F\mathbf{z}_B \\
    \mathbf{J}\dot{\boldsymbol{\omega}} &= - \boldsymbol{\omega}\times\mathbf{J}\boldsymbol{\omega} + \mathbf{m}\nonumber 
\end{align}
where $m$ is the UAV mass, $\mathbf{r} = [x, y, z]^\top$ is the UAV position vector in world frame $\mathcal{W}$, $g$ is the gravity, $F$ is the body thrust, $\mathbf{J}$ is the moment of inertia along the body principle axes. $\boldsymbol{\omega}$ is the robot angular velocity, $\mathbf{m} = [m_x, m_y, m_z]$ is the body moment. We use 3-2-1 Euler angles $\boldsymbol{\Omega} = [\phi, \theta, \psi]$ to define the orientation.
The system state is defined as: $\mathbf{x} = [\mathbf{r}, \boldsymbol{\Omega}, \dot{\mathbf{r}}, \boldsymbol{\omega}]^\top$, and control input $\mathbf{u} = [F; \mathbf{m}]^\top$.

\subsubsection{Trajectory Generation} It has been proved that the quadrotor dynamics is \textit{differential flat} with the $\textit{flat outputs}$ as $\boldsymbol{\sigma} = [\mathbf{r}^\top, \psi]^\top \in \mathbb{R}^4 ~$\cite{Mellinger2011MinSnap}. In other words, the states and the inputs can be written as algebraic functions of the UAV position and heading and their derivatives. Therefore, the trajectory planning can be simplified as planning for $\boldsymbol{\sigma}$ rather than the full dynamic states, which also ensures the dynamic feasibility. Computation efficiency can thus be improved compared to the traditional nonlinear MPC.
In fact, the inputs $m_x$ and $m_y$ appear as functions of the $4^{\mathrm{th}}$ derivatives of the positions (snap), $m_z$ appears in the $2^{\mathrm{nd}}$ derivative of the yaw angle, the TO problem is formulated as minimizing the total snap and control effort along a time horizon $t_n$ given key-point time stamp as $\mathcal{T} = [t_1, ..., t_n]^\top \in \mathbb{R}^n$:
%We treat each dimension of the flat outputs as a separate TO problem. For instance, the $x$ component of the trajectory, $\boldsymbol{x}(t)$, is represented as a smooth $N=2\epsilon-1$ degree piecewise polynomial. In particular, $\boldsymbol{x}(t)$ consists of $n-1$ sequential segments, denoted as $\boldsymbol{x}_1(\cdot), \boldsymbol{x}_2(\cdot), ...,\boldsymbol{x}_{n-1}(\cdot),$. Each segment $i$ is parameterized by a coefficient vector $\mathbf{c}_i=[c_0, c_1, ..., c_N]$ and a duration $\Delta t = t_{i+1}-t_i$. The overall $\boldsymbol{x}(t)$ is thus fully determined by the coefficient matrix $\mathbf{c} = [\mathbf{c}_1^\top,...,\mathbf{c}_{n-1}^\top]^\top$ and the key-point time stamp $\mathcal{T}$,
% \begin{align}
%     \boldsymbol{x}_i(t) &= \mathbf{c}_i^\top \boldsymbol{\Omega}(t) \notag \\
%     \text{where} \quad \boldsymbol{\Omega}(t) &= [1, t, t^2, ..., t^N]^\top, \forall t \in [t_i, t_{i+1}].
% \end{align}
% The $y$, $z$, and $\psi$ components of the trajectory can also be defined in similar manner. Then, we construct a QP problem to minimize the control effort (snap, $\epsilon=4$) for executing the planned time-parametrized trajectory.
\begin{subequations}\label{pro:opt}
\begin{align}
    \min &\int_{0}^{t_n}\mu_r \left\lVert\frac{d^4\mathbf{r}}{dt^4}\right\rVert^2 + \mu_{\psi} \left(\frac{d^2\psi}{dt^2}\right)^2dt  
    \label{opt_obj}\\
    \text{s.t.}\ \  &\mathbf{r}^{[3]}(0) = \bar{\mathbf{r}}^{[3]}_0, \mathbf{r}^{[3]}(t_n) = \bar{\mathbf{r}}^{[3]}_T, \notag\\
    &\psi^{[1]}(0) = \bar{{\psi}}^{[1]}_0,  \psi^{[1]}(t_n) = \bar{\psi}^{[1]}_T,
    \label{opt_se}\\
    &\boldsymbol{\sigma}_i(t_{i+1}) = \boldsymbol{\xi}_{i+1}
    \label{opt_m}\\
    &\mathbf{r}_i^{[3]}(t_{i+1}) = \mathbf{r}_{i+1}^{[3]}(t_{i+1}), \psi^{[1]}_i(t_{i+1}) = \psi^{[1]}_{i+1}(t_{i+1})
    \label{opt_con}
\end{align}
\end{subequations}
where $\mu_r$ and $\mu_{\psi}$ are constant weight factors to make the integral nondimensional. 
$\boldsymbol{\sigma}(t)$ are piecewise polynomials of order $m$ over $n$ (Here $m = 7$) time intervals for each dimension of the flat output $\mathbf{r}(t)$ and $\psi(t)$ parameterized by time $t$. $\boldsymbol{\sigma}_i(t) = [\mathbf{r}_i(t)^\top, \psi_i(t)]^\top$ denotes the polynomials of the $i^{\mathrm{th}}$ segment between time interval $t_i$ and $t_{i+1}$. $(\cdot)^{[k]}$ represents the quantity up to $k^{\mathrm{th}}$ order derivative, e.g $\mathbf{r}^{[k]}(t) = [\mathbf{r}(t)^\top, \mathbf{r}'(t)^\top, ..., \mathbf{r}^{(k)}(t)^\top]^\top$.
$(\cdot)_0, (\cdot)_T$ are initial and final boundary conditions. 
Specifically, \eqref{opt_se} constrains the waypoint and its higher-order derivatives at the start and end points, the intermediate waypoints is constrained by \eqref{opt_m}, while \eqref{opt_con} ensures the continuity of the position and its higher-order derivatives at the intermediate waypoints. 
Notice that the hard constraints can be relaxed to inequality constraints to incorporating flight corridor,  actuator limit, etc.
%Furthermore, to provide greater flexibility of TO, we can relax the hard constraints on the intermediate waypoints by introducing inequality constraints (same manner as \cite{Mellinger2011MinSnap}). This allows the trajectory to be confined within a flight corridor of a certain width. As illustrated in Fig.\ref{fig:corridor}, different corridor widths provide us the freedom of controlling the TO. 

In fact, we can represent $\boldsymbol{\sigma}_i = \mathbf{c}_i^\top \boldsymbol{\Omega}(t)$, where $\boldsymbol{\Omega}(t) = [1, t, ..., t^m]^\top$. With the coefficients of all the piecewise polynomials as the decision variables $\mathbf{c} = [\mathbf{c}_1^\top, ..., \mathbf{c}_{n}^{\top}]^\top$, Problem~(\ref{pro:opt}) can thus be formulated as a QP problem: 
\begin{align}\label{pro:QP}
    \min_{\mathbf{c}} &\ \mathbf{c}^\top \mathbf{Q} \mathbf{c} \\
    \text{s.t.}\ \ &\mathbf{A} \mathbf{c} = \mathbf{b}, \quad\mathbf{G}\mathbf{c} \leq \mathbf{h} \nonumber
\end{align}
where $\mathbf{A}$, $\mathbf{b}$ encapsulate the equality constraints, and $\mathbf{G}$, $\mathbf{h}$ describe the inequality constraints. Notice that the key point path $\boldsymbol{\xi}$ is predicted by the front-end network planner $f(\boldsymbol{\theta}_f)$. Problem~(\ref{pro:QP}) is the inner/lower level optimization of Problem~\eqref{pro:blo}. With optimized $\mathbf{c}^*$, we then evaluate $\boldsymbol{\sigma}(t)$ at control frequency to obtain the planned trajectory $\boldsymbol{\tau}^*$. 

\subsubsection{Time Allocation}
In Problem~(\ref{pro:opt}), the arrival times at different key points are critical but cannot be predetermined, especially in dynamic environments. A common approach in previous work is to solve a separate iterative optimization problem using gradient descent with backtracking line search~\cite{Mellinger2011MinSnap}. However, this will lead to a tri-level optimization in our setup, significantly degrading real-time performance. 
To address this, we develop a Time Allocation Net (TAN) $g(\boldsymbol{\theta_T})$ that predicts the time allocation for each segment of a key-point trajectory. It takes the key-point path $\boldsymbol{\xi}$ as input and output the allocated times $\mathcal{T}$. We use the traditional gradient-descent-based method as in ~\cite{Mellinger2011MinSnap} to compute optimal time allocations for training. Then, during inference, our network efficiently predicts the time allocations in real time. In practice, TAN utilizes an MLP followed by a softmax layer, predicting the time allocation percentage for each trajectory segment relative to the total time.
Notice that TAN was jointly trained with the front-end network with the supervised reference generated by the optimization method~\cite{Mellinger2011MinSnap}, rather than expert demonstration or human annotations.

\subsubsection{Differentiable Optimization}
Since $\boldsymbol{\tau}^*$ is a function of $f(\boldsymbol{\theta}_f)$ and $g(\boldsymbol{\theta}_T)$, and will be passed to the upper level optimization for the final end-to-end training, the key part is to calculate the implicit gradient $\frac{\partial \boldsymbol{\tau}^*}{\partial \xi}$. 
The traditional unrolling approach maintain the computational graph throughout the entire iteration process, which poses significant computation burden. It may also run into divergent or gradient vanishing. In this work, we employ the implicit function differentiation theorem and leveraging the KKT condition of the Problem~(\ref{pro:QP}) at the optimal point to analytically compute the gradients of the parameters. Thus, there is no need for explicit unrolling of the entire iteration process.
In practice, we leverage a fast differentiable QP solver for PyTorch as the optimization layer~\cite{cvxpylayers2019, amos2017optnet}.

\subsection{3D Cost Map and Training Loss}

\begin{figure}
    \centering
    \subfigure[Point Cloud]{
        \includegraphics[width=0.17\textwidth]{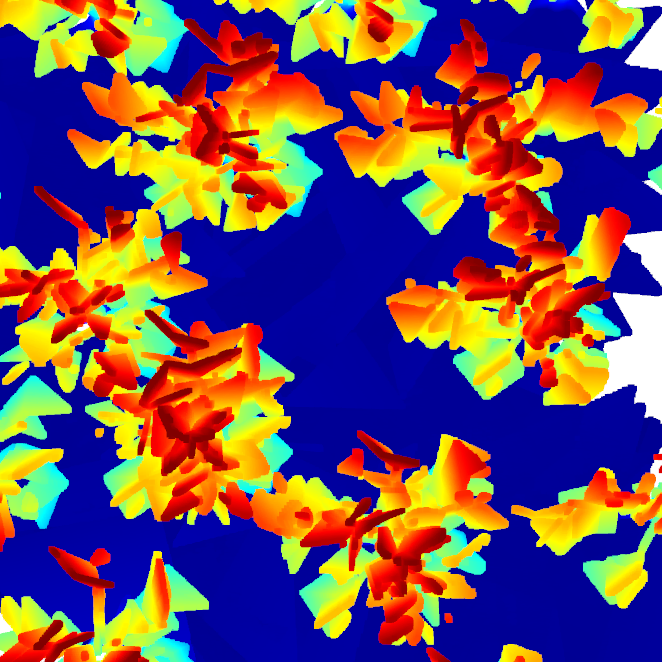}
    }\hspace{-2mm}
    \subfigure[ESDF at h=1.0m]{
        \includegraphics[width=0.17\textwidth]{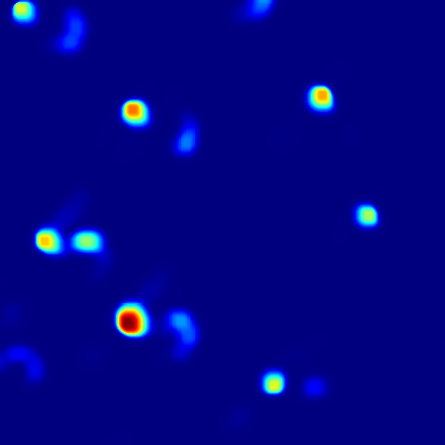}
    }\hspace{-2mm}\vspace{-2mm}
    
    \subfigure[ESDF at h=2.5m]{
        \includegraphics[width=0.17\textwidth]{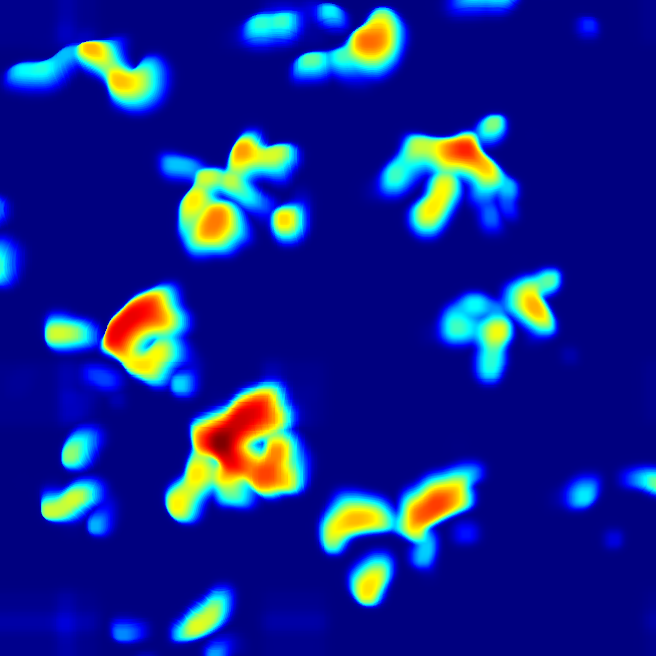}
    }\hspace{-2mm}
    \subfigure[ESDF at h=3.5m]{
        \includegraphics[width=0.17\textwidth]{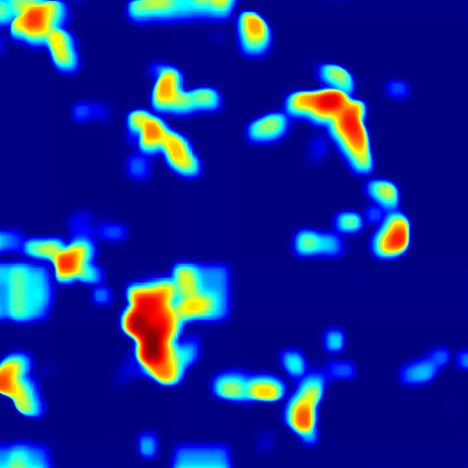}
    }\hspace{-2mm}
    \caption{3D ESDF cost map in the forest environment. (a) Reconstructed point cloud from depth images. (b–d) ESDF slices at different altitudes, color-coded by cost.} %(b), (c), and (d) showcase the 3D ESDF. We slice the ESDF map at different altitudes and color code the cost of the corresponding points.}
    \label{fig: esdf}
    \vspace{-2em}
\end{figure}

\subsubsection{3D Cost Map} %We develop a 3D cost map that provides traversability costs to guide UAV behavior. Based on the collected depth images, UAV poses, and the associated camera poses, we reconstruct the environment offline before the training. 
We develop a 3D cost map that provides collision costs to guide UAV behavior by first reconstructing the environment offline from depth images, UAV poses, and camera poses. 
A 3D Euclidean Signed Distance Field (ESDF) like Figure \ref{fig: esdf} is then built based on the reconstructed environment. Unlike common approaches that assign distances only in obstacle regions (leaving free space as a constant), we also label the distance in free space to its nearest obstacle boundary. This ensures existence of valid gradient for network learning, as large internal volume of free space in the 3D space would otherwise cause gradient vanishing. Finally, we smooth the ESDF with a Gaussian filter to improve local differentiability. This 3D cost map enables self-supervision for our BLO framework.
%In particular, unlike the common approaches that use ESDF for path planing by only computing the distance to the nearest free (non-obstacle) space for each position in the obstacle space while keep the free space constant, we also label the distance to the nearest boundary of the obstacle in the free space to ensure the existence of valid gradient for network learning. Because the network could be more sensitive to the gradient vanishing in the 3D space due to the large internal volume of free space by predicting paths through which will completely lose the learning ability if the ESDF are all constants. The ESDF is then smoothed with a Gaussian filter to get local differentiability. %A 3D cost map $\mathcal{H}$ with non-negative cost at both free and obstacle space is created. 

\begin{figure*}
    \centering
    \subfigure[Office]{
        \includegraphics[width=5.0cm]{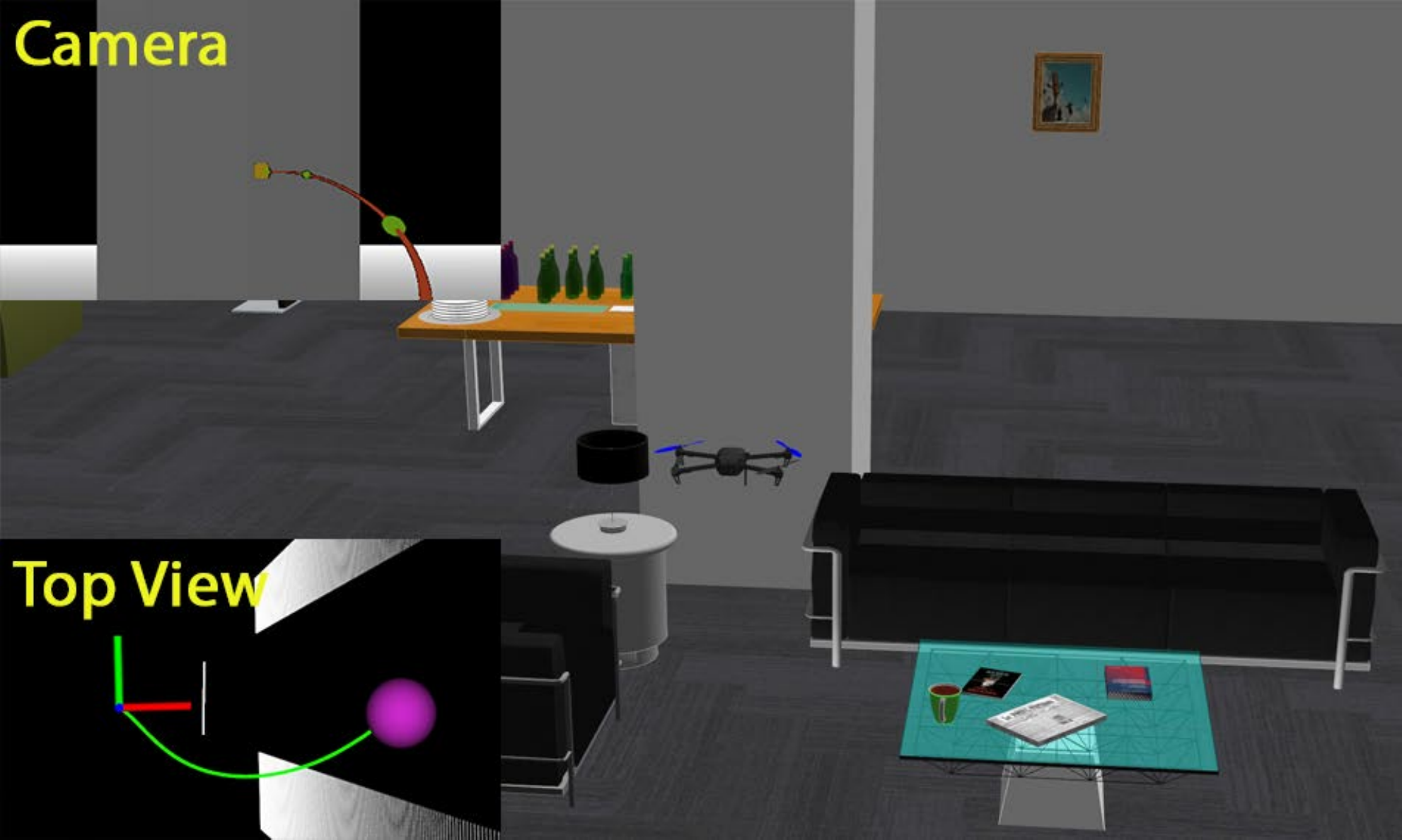}
        \label{fig:office_stack}
    }\hspace{-3mm}
    \subfigure[Garage]{
        \includegraphics[width=5.0cm]{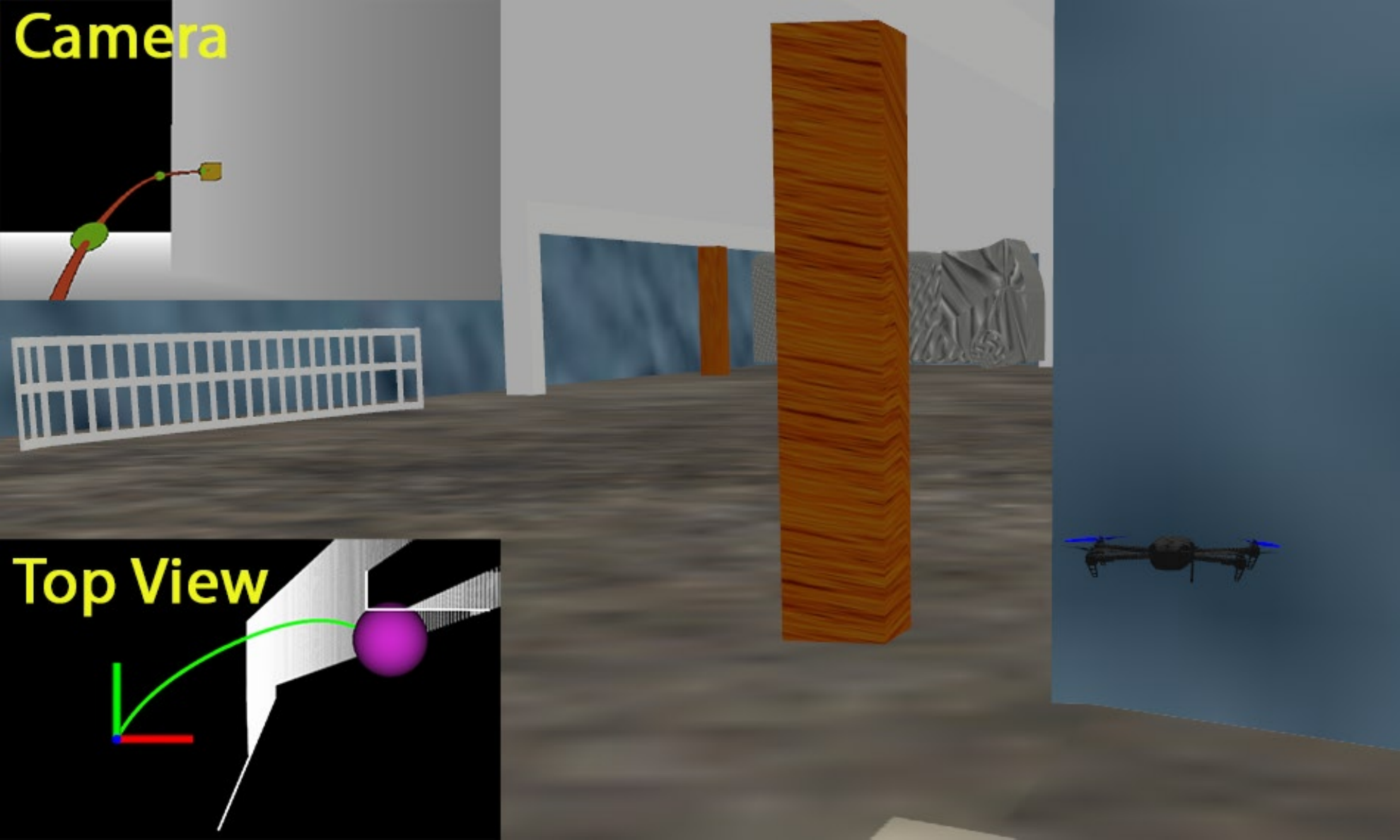}
        \label{fig:garage_stack}
    }\hspace{-3mm}
    \subfigure[Forest]{
        \includegraphics[width=5.0cm]{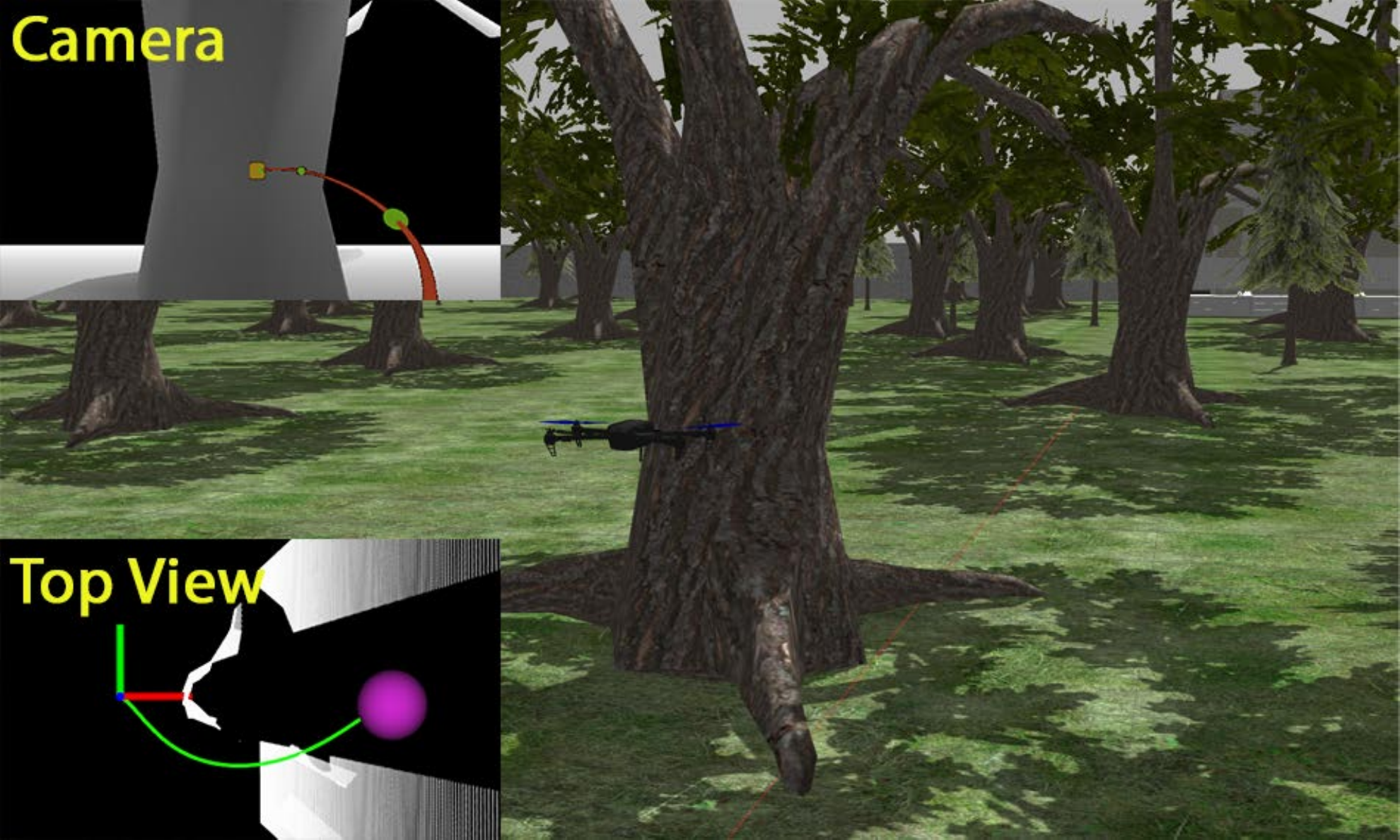}
        \label{fig:forest_stack}
    }\hspace{-3mm}\vspace{-3mm}
    \caption{Illustration of different simulation environments. The purple spheres represent goal points, while the green curve indicates the planned trajectories.}
    \label{fig: sim_env}
    \vspace{-2em}
\end{figure*}

\subsubsection{Training Loss}
Based on the 3D cost map, the upper-level training loss is defined as the weighted summation of obstacle cost $\mathcal{U}^{\mathcal{O}}$, target cost $\mathcal{U}^{\mathcal{G}}$, smoothness cost $\mathcal{U}^{\mathcal{S}}$, and escape cost $\mathcal{U}^{\mathcal{E}}$ in a similar way as~\cite{Yang-RSS-23}. We also add the time allocation cost $\mathcal{U}^{\mathcal{T}}$, e.g.:
\begin{align}
    \mathcal{U}(\boldsymbol{\tau}, \boldsymbol{\theta}_T) = &\gamma_1 \mathcal{U}^{\mathcal{O}}(\boldsymbol{\tau}) + \gamma_2 \mathcal{U}^{\mathcal{G}}(\boldsymbol{\tau}) + \gamma_3 \mathcal{U}^{\mathcal{S}}(\boldsymbol{\tau}, \mathbf{p}^R, \mathbf{G}) \nonumber\\
    &+ \gamma_4\mathcal{U}^{\mathcal{E}}(\boldsymbol{\tau}) + \gamma_5\mathcal{U}^{\mathcal{T}}(g(\boldsymbol{\theta}_T))
\end{align}
where $\gamma_k \in \mathbb{R}^+, k = 1,2,...,5$ are weight factors for different terms.
Obstacle cost is defined as
\begin{equation}
    \mathcal{U}^{\mathcal{O}}(\boldsymbol{\tau}) = \sum_i\text{ESDF}(\mathbf{p}_i), \quad \mathbf{p}_i\in \boldsymbol{\tau} 
\end{equation}
where each point $\mathbf{p}_i$ on trajectory $\boldsymbol{\tau}$ will be projected on our 3D ESDF map to get the cost value $\text{ESDF}(\mathbf{p}_i)$.

Target cost is the Euclidean distance from the final point $\mathbf{p}_f$ on $\boldsymbol{\tau}$ to the target $\mathbf{G}$, e.g. 
\begin{equation}
    \mathcal{U}^{\mathcal{G}}(\boldsymbol{\tau}) = \|\mathbf{p}_f - \mathbf{G}\|_2
\end{equation}

Smoothness cost is defined as the deviation between the segment-wise distances of the generated trajectory and those of a direct linear connection between the start and goal positions:
% \begin{equation}
%      \mathcal{C}^{\mathcal{S}}(\boldsymbol{\sigma}, \mathbf{p}^R, \mathbf{G}) = \sum_{i=0}^{n-2} \left\lvert \frac{d(\mathbf{p}^R, \mathbf{G})}{n-1} - d_{\boldsymbol{\sigma}}(\mathbf{\xi}_i, \mathbf{\xi}_{i+1})\right\rvert
% \end{equation}
\begin{equation}
\mathcal{U}^{\mathcal{S}}(\boldsymbol{\tau}) = \sum_{i=1}^{N-1} \left| \|\mathbf{p}_{i+1} - \mathbf{p}_{i}\|_2 - \|\mathbf{\hat{p}}_{i+1} - \mathbf{\hat{p}}_{i}\|_2 \right|
\end{equation}
where $N$ is the number of waypoints, $\mathbf{p}_{i}$ is $i^{\mathrm{th}}$ waypoint of predicted trajectory,  $\mathbf{\hat{p}}_{i}$ denotes the $i^{\mathrm{th}}$ waypoint of the straight line connecting start point and goal point. It ensures evenly distributed length of each trajectory segments between consecutive waypoints. By aligning the segment lengths more closely with those of a straight-line path, the optimization process reduces unnecessary oscillations and ensures a more smooth motion.

Escape loss is to provide the planner with the flexibility to escape from the local minima. Rather than setting large obstacle costs which could cause over-conservative policy, we define the escape loss as
\begin{equation}
    \mathcal{U}^{\mathcal{E}}(\boldsymbol{\tau}) = \left\{ \begin{array}{ll}
       \text{BCELoss}(\eta, 1.0)  &  \boldsymbol{\tau} \bigcap \mathcal{O} \neq \varnothing\\
       \text{BCELoss}(\eta, 0.0)  &  \text{otherwise}.
    \end{array}\right.
\end{equation}
where $\eta$ is a network predicted collision probability for each trajectory. During deployment, the planner will execute the trajectory with $\eta < 0.5$. BCE is the binary cross entropy.

Time allocation cost guides the update of Time Allocation Net during training. Specifically, it is trained to minimize the discrepancy between its predicted time allocation and the optimal time allocation $\mathcal{T}^*$ obtained via iterative optimization with backtracking line search.

\begin{equation}
\mathcal{U}^{\mathcal{T}}(g(\boldsymbol{\theta}_T)) = \frac{1}{n} \sum_{i=1}^{n} \left\| g(\boldsymbol{\theta}_T)_i - \mathcal{T}^*_i \right\|^2
\end{equation}

\subsection{Bi-level Optimization}
The whole pipeline forms a BLO problem, where the differentiable MSTO is the lower-level optimization taking the output of the planner network and generate the trajectory $\boldsymbol{\tau}^*(f(\boldsymbol{\theta}_f), g(\boldsymbol{\theta}_T))$ to optimize the control effort $\mathcal{L}$. Then, the upper-level optimization finds the network parameter $\boldsymbol{\theta}_f$ and $\boldsymbol{\theta}_T$ to optimize the overall training loss $\mathcal{U}$. Thus, the gradient of the training loss $\mathcal{U}$ must be propagated back through the lower-level optimization, and then the network parameters can be updated using gradient descent:
%To update the network parameter using gradient decent, the gradient of the training loss $\mathcal{U}$ with respect to the network parameter $\boldsymbol{\theta}_f$ and $\boldsymbol{\theta}_T$ can be computed as:
\begin{align}
    \nabla_{\theta_f}\mathcal{U} &= \frac{\partial \mathcal{U}}{\partial\boldsymbol{\tau}^*}\frac{\partial \boldsymbol{\tau}^*}{\partial g} \frac{\partial g}{\partial f} \frac{\partial f}{\partial \boldsymbol{\theta}_f} + \frac{\partial \mathcal{U}}{\partial \boldsymbol{\tau}^*}\frac{\partial\boldsymbol{\tau}^*}{\partial f} \frac{\partial f}{\partial \boldsymbol{\theta}_f}\\
    \nabla_{\theta_T}\mathcal{U} &= \frac{\partial \mathcal{U}}{\partial\boldsymbol{\tau}^*}\frac{\partial \boldsymbol{\tau}^*}{\partial g} \frac{\partial g}{\partial \boldsymbol{\theta}_T} +  \frac{\partial \mathcal{U}}{\partial g}\frac{\partial g}{\partial \boldsymbol{\theta}_T} \\
    \boldsymbol{\theta}_{f, t+1} &= \boldsymbol{\theta}_{f, t} -\alpha\cdot\nabla_{\theta_f}\mathcal{U}\\ \boldsymbol{\theta}_{T, t+1} &= \boldsymbol{\theta}_{T, t} -\alpha\cdot\nabla_{\theta_T}\mathcal{U}
\end{align}
where $\alpha$ is the learning rate during training. Usually, $\frac{\partial \mathcal{U}}{\partial \boldsymbol{\tau}^*}$ and $\frac{\partial \mathcal{U}}{\partial g}$ can be explicitly computed with the expression of $\mathcal{U}$. $\frac{\partial f}{\partial \boldsymbol{\theta}_f}$ and $\frac{\partial g}{\partial \boldsymbol{\theta}_T}$ can also be easily obtained from the computation graph of the network forward pass. The challenging part is to compute $\frac{\partial \boldsymbol{\tau}^*}{\partial \boldsymbol{\theta}_f}$ and $\frac{\partial \boldsymbol{\tau}^*}{\partial \boldsymbol{\theta}_T}$ due to the $\mathrm{argmin}$ operation. Thanks to our differentiable optimization developed in Sec.~\ref{sec:diff-opt}, both of them can be computed without the need of unrolling the entire iteration process of the optimization.

\section{Experiments}
\label{sec: experments}
We conduct experiments in both simulated and real-world environments to assess the effectiveness of our method. We evaluate navigation tasks in various settings and compare our approach to several baseline methods.

\subsubsection{Experimental Settings}
For simulation, we use the open-source Autonomous Exploration Development Environment~\cite{cao2022autonomous} and our customized UAV Gazebo simulator, running on a 2.4GHz i9 laptop with an NVIDIA RTX 4060 GPU. Real-world experiments are conducted on our custom-built quadrotor UAV with an NVIDIA Jetson Orin onboard computer running the planning pipeline and a Pixracer autopilot running our customized PX4 firmware. 
At runtime, trajectory optimizer outputs a time-parameterized sequence (position, velocity, acceleration) as the onboard outer-loop control reference.
A front facing Intel RealSense D435 camera provides depth perception at 30 Hz. A motion capture system is used for indoor localization, which can be easily switched to visual inertial odometry \cite{QINTONG2018VINSMONO,QINTONG2018VINSFUSION} for outdoor navigation. %Since our planner plans dynamically feasible trajectories incorporating position, velocity, and acceleration (p, v, a) simultaneously, we directly send (p, v, a) commands to our customized outer loop control stack.

\subsubsection{Training Data}
We collected training data from both simulated and real-world environments. In simulation, we used a joystick to manually fly the UAV in Gazebo simulator, collecting 3 datasets from the environments in Figure \ref{fig: sim_env} (office, garage, and forest), which contains approximately 2000 depth images and corresponding robot poses. We also collected 7 datasets in real-world to account for perception noise and varying lighting conditions. Notice that our method does not require labeled ground-truth data or human expert supervision; its training is fully self-supervised, relying solely on a well-designed loss and the 3D cost map. 

\subsubsection{Training Details}
The network follows an encoder–decoder architecture. A ResNet-18 backbone encodes visual observations, while the planning module fuses the encoded features with a linearly projected goal embedding. The fused representation is processed by two convolutional layers and a three-layer MLP to generate $k$ 3D waypoints, with a parallel MLP branch predicting a confidence score. The networks are trained with batch size of 16 using the Adam optimizer with a learning rate 0.0001. We used the pre-trained model from iPlanner~\cite{Yang-RSS-23} and trained it for 50 epochs on an NVIDIA 4060 GPU. The entire training process took approximately 70 hours.

\subsubsection{Baselines}
For the non-learning method, we choose two traditional approaches as baselines. \textbf{MP}~\cite{zhang2020MP}: Generate trajectories by searching over a set of pre-defined dynamically feasible motion segments. \textbf{EGO-Planner}~\cite{zhou2020ego}: A gradient-based planner without the ESDF construction. For the learning-based method, we adopt \textbf{iPlanner}~\cite{Yang-RSS-23} as the baseline, which has already compared with other learning-based methods, including RL. Since iPlanner only plans for 2D motion, we re-train it on our 3D cost map. All methods use a front-facing stereo vision depth camera. 
%, which also serves as a performance reference for learning-based methods
%\subsubsection{Evaluation Metrics}
\vspace{-0.5em}
\subsection{Simulation Experiments}
\begin{figure}
    \vspace{1mm}
    \centering
    \includegraphics[width=1\linewidth]{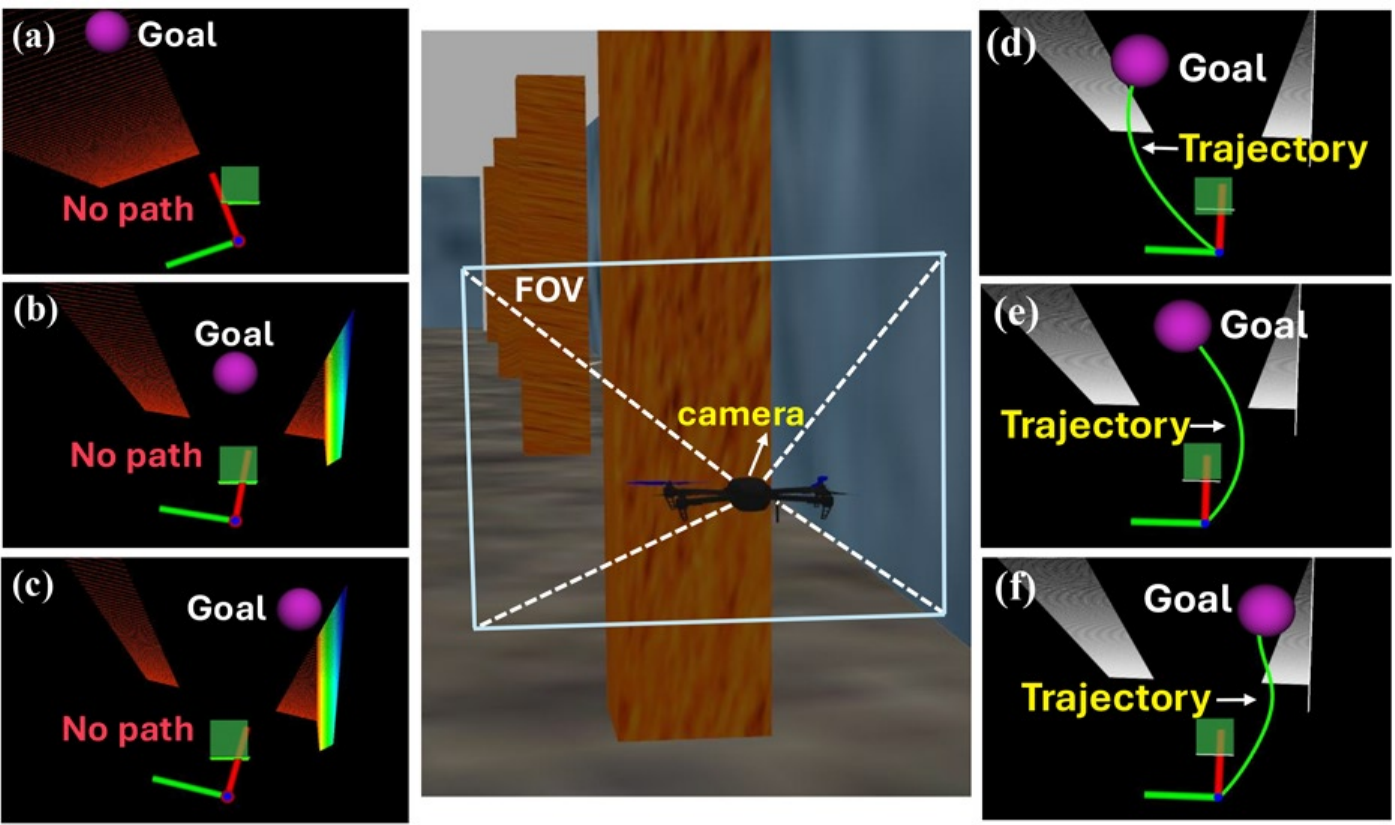}
    \caption{Navigation in narrow space. (a)(b)(c) MP approach gets stuck in local minima, failing to generate feasible trajectories. (d)(e)(f) Our method is more robust to viewpoint variations, successfully planning trajectories regardless of whether the goal is on left, center, or right behind the obstacle.}
    \label{fig:local_minima}
    \vspace{-2em}
\end{figure}
%(d)–(f): Our method is more robust to viewpoint variations, successfully planning trajectories whether the goal is to the left, center, or right behind the obstacle.
\subsubsection{Overall Performance} We perform simulation experiments in Gazebo as shown in Figure \ref{fig: sim_env}. To evaluate the overall planning performance, we first measure the success rate, defined as the UAV reaching its goal from its starting point without any collision along the flight path. Specifically, we sample 60 start-goal pairs per environment. As shown in Table \ref{table:success_rate}, our method outperforms iPlanner in all cases because our differentiable MSTO module accounts for UAV dynamics in 3D, producing dynamically feasible trajectories. Although the traditional MP performs slightly better in some scenarios, it can get stuck in local minima from which it cannot escape. For example, as shown in Figure \ref{fig:local_minima}, when the vehicle moves directly behind a pillar, its field of view is heavily restricted. In this scenario, target points located within a large area behind the obstacle often lead the MP planner to local minima. In contrast, our method demonstrates more robust, successfully generating collision-free and feasible trajectories. We notice that EGO-Planner's success rate is close to 100\% under all environments, our method has more benefits on control effort (will discuss more in next section). 

\begin{table}[h!]
    % \vspace{1.55mm}
    \centering
    \caption{Success Rate (\%) ($\uparrow$)}
    \vspace{-10pt}
    
    \begin{tabular}{c|cccc}
    \toprule
    Method & Office & Garage & Forest & Overall \\
    \midrule
    MP & 86.7 & \textbf{96.7} & 48.3 & 77.2 \\
    iPlanner & 75.0 & 78.3 & 63.3 & 72.2 \\
    Ours & \textbf{96.7} & 91.7 & \textbf{76.7} & \textbf{88.3} \\
    \bottomrule
    \end{tabular}
    
    \label{table:success_rate}
    \vspace{-1.0em}
\end{table}

\begin{figure}
    \centering
    \subfigure[No corridor constraints]{
        \includegraphics[width=0.42\linewidth]{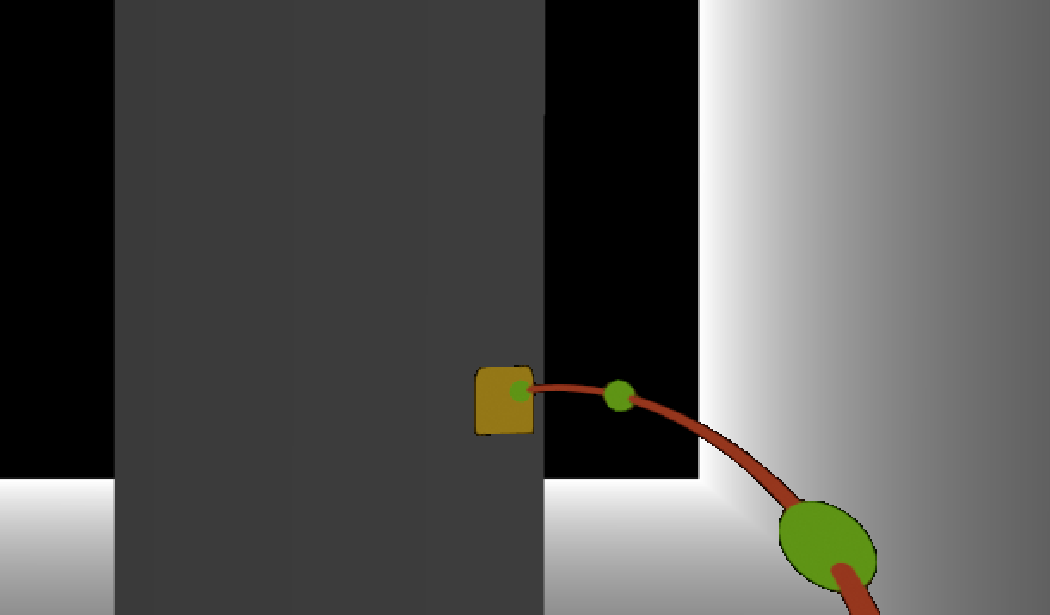}
        \label{fig:corridor1}
    }\hspace{-2mm}
    \subfigure[Corridor width $l_c=0.05$m]{
        \includegraphics[width=0.42\linewidth]{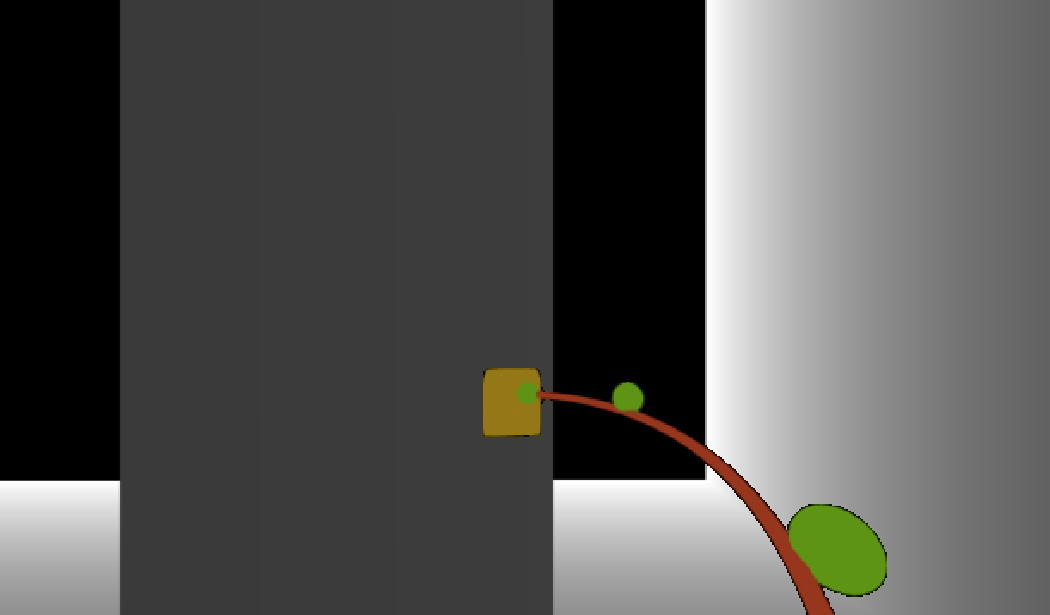}
        \label{fig:corridor2}
    }\vspace{-2mm}
    \subfigure[Corridor width $l_c=0.10$m]{
        \includegraphics[width=0.42\linewidth]{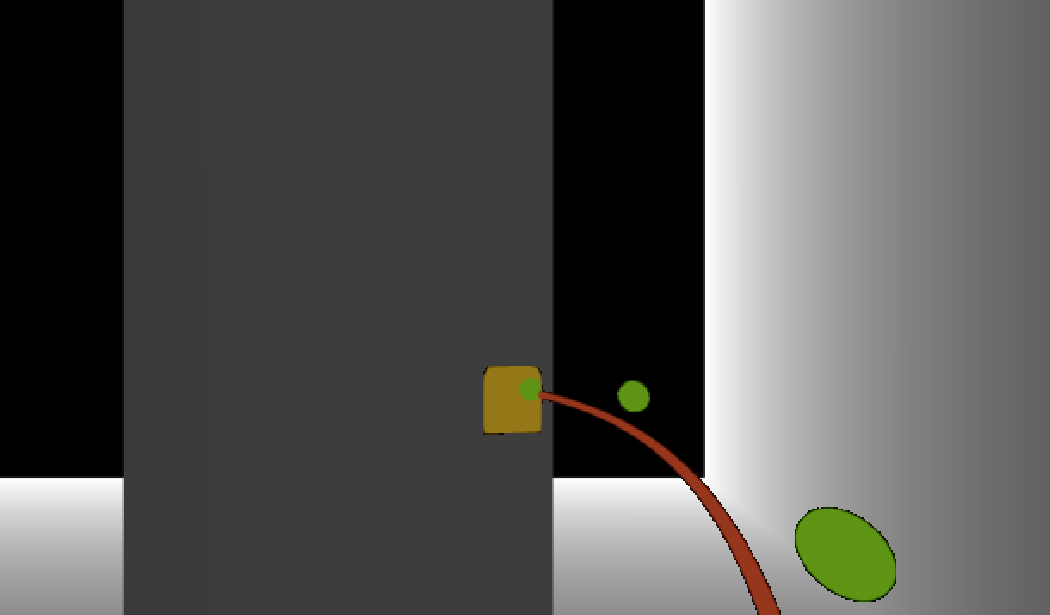}
        \label{fig:corridor3}
    }\hspace{-2mm}
    \subfigure[Corridor width $l_c=0.20$m]{
        \includegraphics[width=0.42\linewidth]{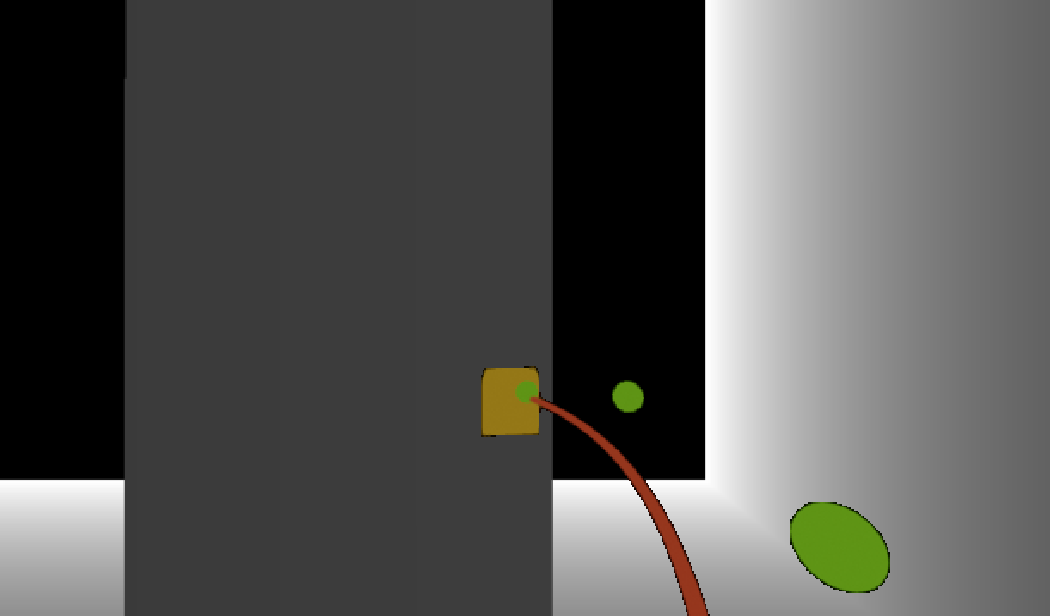}
    }\vspace{-2mm}
    \caption{Planning performance when incorporating corridor constraints. The optimized trajectories deviate from the planned key-point path to satisfy the corridor (inequality) constraints.}
    \label{fig:corridor}
    \vspace{-2em}
\end{figure}

\subsubsection{Control Effort} We then evaluate and compare the control effort of different methods through the low-level optimization cost $\mathcal{L}$ (integral of the squared Snap) in our differentiable MSTO module, see Table \ref{table:planning_latency}. Thanks to the deliberate minimization of the snap of 3D trajectory, our method achieves the lowest total snap, demonstrating its superiority in UAV planning over other baseline approaches. %Here control effort is defined as the integral of the squared Snap.
    
\begin{table}[h!]
    \vspace{-0.5em}
    \centering
    \caption{Control Efforts and Planning Latency}
    \vspace{-7pt}
    
    \begin{tabular}{c|cc|cc}
    \toprule
    \multirow{2}{*}{Method} & \multicolumn{2}{c|}{ Control Effort (m$^2$/s$^7$) ($\downarrow$)} & \multicolumn{2}{c}{Latency (ms) ($\downarrow$)} \\
    & Mean & Std & Mean & Std \\
    \midrule
    MP & 97.65 & 32.52 & 29.13 & 4.20 \\
    Ego & 32.97 & 7.83 & 20.08 & 3.15 \\
    iPlanner & 58.24 & 11.95 & \textbf{7.51} & \textbf{0.97} \\
    Ours & \textbf{21.16} & \textbf{4.86} & 13.16 & 2.28 \\
    \bottomrule
    \end{tabular}
    
    \label{table:planning_latency}
    \vspace{-1em}
\end{table}

\begin{figure*}
    \vspace{1em}
    \centering  
    \includegraphics[width=1\textwidth]{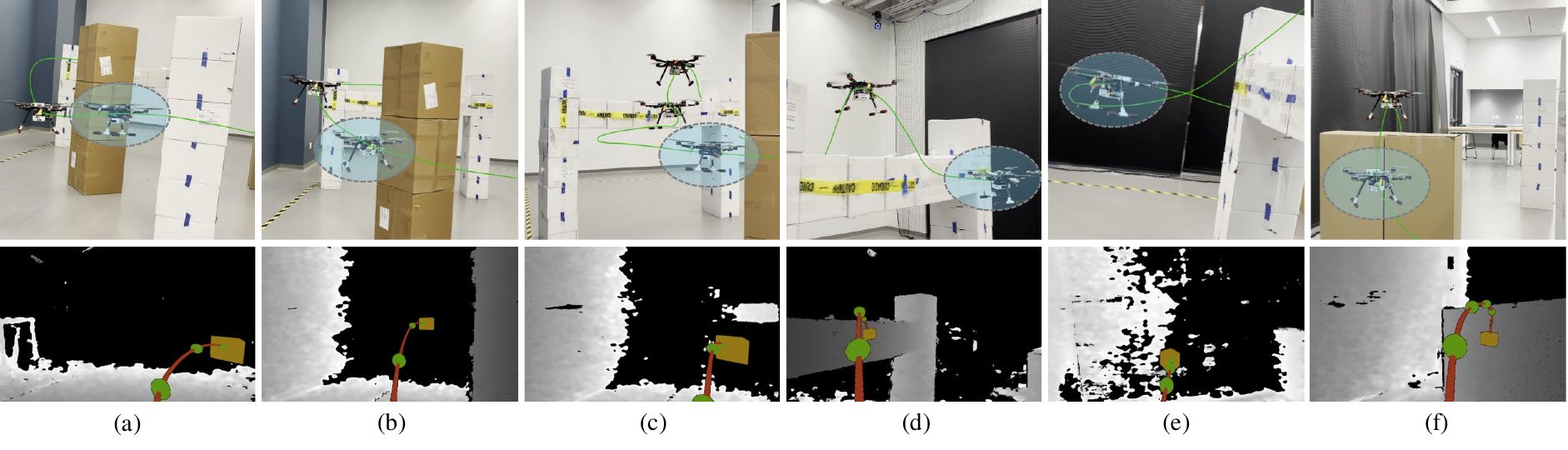}
    \vspace{-2em}
    \caption{A sequence of snapshots of real-world flight experiments in complex environment. The depth images are captured from the onboard camera of the UAV in blue circles. The green curve showcases the UAV's trajectory. (a-c) The UAV avoids multiple vertical pillars;  then (d-e) turns around and avoids a horizontal beam; finally (f) passes another obstacle with different height.}
    \label{fig: exp_snapshot}
    \vspace{-1.5em}
\end{figure*}

%D UAV trajectory planning in complex environment. Different desired waypoints are given by user, the UAV relies solely on depth images as input. The green curve showcases the UAV's trajectory. Starting from point (A) avoids vertical pillar (B), then performs maneuver to position (C) avoids the horizontal beam, (D) performs another vertical avoidance maneuver at different height and reach the blue target point.

\subsubsection{Computational Efficiency} Moreover, we compare the computational efficiency of different methods, as shown in Table \ref{table:planning_latency}. The MP planner incurs significant higher planning latency due to its complex, modularized pipeline. Although iPlanner achieves the lowest latency by relying on a closed-form solution for lower-level trajectory optimization without iterative steps, it thereby loses the ability to enforce physical constraints. In contrast, our method uses iterative optimization for the differentiable MSTO with proper gradient backpropagation, yet still achieves competitive low latency. 

\subsubsection{Inequality Constraints} Thanks to the iterative optimization in our MSTO, we can explicitly incorporate inequality constraints, such as flight corridor or actuator limit. Figure \ref{fig:corridor} shows a scenario where we incorporate corridor constraints. Our method successfully optimizes the planned path within different required corridors. This capability is crucial for agile flight in constrained environments.

\subsubsection{Time Allocation} We finally evaluate our TAN by comparing different time allocation strategies: uniform, acceleration–deceleration, and gradient descent with line search \cite{Mellinger2011MinSnap}, see Table~\ref{table:timeallocation}. We can see that the uniform strategy is simple but ignores dynamic feasibility and environmental constraints, performing poorly in complex settings. The Acceleration-Deceleration strategy, usually built on a 5th-order polynomial, ensures smooth acceleration and deceleration with multiple initial and terminal conditions but may fail to optimize time allocation in obstacle-laden or dynamically constrained environments. Gradient Descent with line search cannot satisfy the real-time efficiency and is highly sensitive to the initial guess. In contrast, our approach incorporates iterative line search as the supervision to address these challenges, achieving optimality and computational efficiency for real-time execution.

\begin{table}[h!]
    \vspace{-1em}
    \centering
    \caption{Performance of Different Time Allocation Strategies}
    \vspace{-7pt}
    
    \begin{tabular}{c|cc|cc}
    \toprule
    \multirow{2}{*}{Method} & \multicolumn{2}{c|}{ Control Effort (m$^2$/s$^7$) ($\downarrow$)} & \multicolumn{2}{c}{Latency (ms) ($\downarrow$)} \\
    & Mean & Std & Mean & Std \\
    \midrule
    Uniform & 601.50 & 131.69 & 13.13 & 2.24 \\
    5th Order Poly & 22.93 & 3.03 & \textbf{11.78} & \textbf{1.82} \\
    Gradient Descent & 26.99 & \textbf{2.59} & 118.19 & 24.58 \\
    Ours & \textbf{21.16} & 4.86 & 13.16 & 2.28 \\
    \bottomrule
    \end{tabular}
    
    \label{table:timeallocation}
    \vspace{-2em}
\end{table}

\subsection{Real-World Experiment}

To validate the real-time effectiveness of our method, we conducted real-world flight experiments in a different physical room from where the training data was collected, with the depth camera as the only perception. The environment involves multiple vertical pillars, walls, horizontal beams and stacked boxes of varying height. Several narrow corridors are naturally generated between these obstacles. The depth measurements are also noisy during flight. Despite these challenges, our UAV successfully performed continuous obstacle avoidance, see Figure \ref{fig: exp_snapshot}. The UAV exhibits smooth left-right and up-down evasive actions to avoid obstacles while maintaining stable trajectories. These results highlight the flexibility and robustness of our approach for obstacle avoidance in constrained 3D environments.

\begin{figure}
    \centering
    \includegraphics[width=0.45\textwidth]{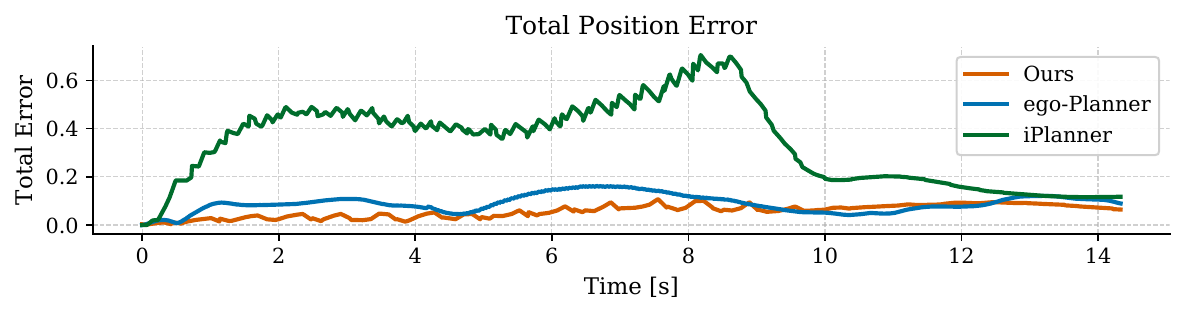}
    \vspace{-1em}
    \caption{Trajectory tracking error in real-world experiment}
    \vspace{-2em}
    \label{fig:tracking_error}
\end{figure}

We also evaluate the overall performance of our approach when avoiding obstacles. 
Table~\ref{table:tracking_error_snap} shows the quantitative evaluation. Because all methods were deployed with the same low-level controller, tracking error directly reflects planning quality. Thanks to the MSTO, our approach achieves the lowest control effort with 27.93 m$^2$/s$^7$ while maintaining competitive tracking accuracy with the mean of 0.0607 m and maximum of 0.1268 m. Figure \ref{fig:tracking_error} illustrates a representative tracking run across all three methods.

\begin{table} [h]
    \vspace{-0.5em}
    \centering
    \caption{Tracking Error and Control Efforts ($\downarrow$)}
    \vspace{-10pt}
    
    \begin{tabular}{c|cc|ccc}
    \toprule
    \multirow{2}{*}{Method} & \multicolumn{2}{c|}{ Control Effort (m$^2$/s$^7$) ($\downarrow$)} & \multicolumn{3}{c}{Tracking Error (m) ($\downarrow$)} \\
    & Mean & Std & Mean & Std & Max \\
    \midrule
    Ego & 40.42 & 18.24 & 0.0910 & 0.0358 & 0.1625 \\
    iPlanner & 55.21 & 13.45 & 0.3422 & 0.1699 & 0.7068 \\
    Ours & \textbf{27.93} & \textbf{8.67} & \textbf{0.0564} & \textbf{0.0269} & \textbf{0.1078} \\
    \bottomrule
    \end{tabular}
    
    \label{table:tracking_error_snap}
    \vspace{-2em}
\end{table}

% Without report snap
% \begin{table} [h]
%     \vspace{-1em}
%     \centering
%     \caption{Tracking Error (m) ($\downarrow$)}
%     \vspace{-10pt}
    
%     \begin{tabular}{c|ccc}
%     \toprule
%     Method & Mean & Std & Max \\
%     \midrule
%     iPlanner & 0.2343 & 0.2013 & 0.6842 \\
%     Ours & \textbf{0.0541} & \textbf{0.0258} & \textbf{0.1294} \\
%     \bottomrule
%     \end{tabular}
    
%     \label{table:tracking_error_snap}
%     \vspace{-1.5em}
% \end{table}

% \begin{table} [h]
%     \vspace{-1em}
%     \centering
%     \caption{Tracking Error ($\downarrow$)}
%     \vspace{-10pt}
    
%     \begin{tabular}{c|ccc}
%     \toprule
%     \multirow{2}{*}{Method} & \multicolumn{2}{c|}{ Control Effort (m$^2$/s$^7$) ($\downarrow$)} & \multicolumn{3}{c}{Tracking Error (m) ($\downarrow$)} \\
%     & Mean & Std & Mean & Std & Max \\
%     \midrule
%     iPlanner & 0 & 0 & 0.2343 & 0.2013 & 0.6842 \\
%     Ours & \textbf{0} & 0 & \textbf{0.0541} & \textbf{0.0258} & \textbf{0.1294} \\
%     \bottomrule
%     \end{tabular}
    
%     \label{table:tracking_error_snap}
%     \vspace{-2em}
% \end{table}
\section{Conclusion}
\label{sec: conclusion}
This paper develops a self-supervised UAV path planning pipeline that integrates a learning-based depth perception with differentiable trajectory optimization. A 3D cost map was introduced to self-supervise UAV behavior. Additionally, we designed a differentiable minimum snap trajectory optimization module to ensure dynamically feasible paths. A time allocation network improves the real-time efficiency and optimality. Our approach thus improves generalizability and interpretability. Both simulation and real-world experiments demonstrate that our method can enable UAV navigate effectively by avoiding obstacles and execute dynamics feasible trajectory across various environments. Our method achieves 30.90\% reduction in control effort compared to the state-of-the-art. Future work includes further testing under diverse operating conditions including dynamic obstacles and degraded lighting conditions. 

% \section*{Acknowledgement}
% We thank Xiangfu Li, Zemu Zhang and Xiangqin Chen for their assistance of hardware platform.

\bibliographystyle{IEEEtran}
\bibliography{root}

\end{document}